\documentclass[conference]{IEEEtran}
\IEEEoverridecommandlockouts
\usepackage{cite}
\usepackage{amsmath,amssymb,amsfonts}
\usepackage{algorithmic}
\usepackage{graphicx}
\usepackage{textcomp}
\usepackage{xcolor}
\usepackage{float}
\usepackage{fancyhdr} 
\usepackage{algorithm}
\usepackage{algorithmic}
\usepackage{dblfloatfix}
\makeatletter
\newcommand\fs@norules{\def\@fs@cfont{\bfseries}\let\@fs@capt\floatc@ruled
	\def\@fs@pre{}%
	\def\@fs@post{}%
	\def\@fs@mid{\kern3pt}%
	\let\@fs@iftopcapt\iftrue}
\makeatother
\floatstyle{norules}
\restylefloat{algorithm}\usepackage{algorithm}
\usepackage{algorithmic}

\fancypagestyle{firstpage}{
	\fancyhf{} 
} 
\def\BibTeX{{\rm B\kern-.05em{\sc i\kern-.025em b}\kern-.08em
    T\kern-.1667em\lower.7ex\hbox{E}\kern-.125emX}}
\begin{document}


\title{A Leakage-Free Stacked Ensemble Method for  Multiclass Classification } 



\author{
	\textbf{S P Sharmila$^{1,2,*}$, and Aruna Tiwari$^{1}$}\\
	\small$^{1}$ Indian Institute of Technology Indore, Madhya Pradesh, India. \\
	$^{2}$Siddaganga Institute of Technology, Tumakuru, Karnataka, India.\\
	E-mail: \{phd2201101012, artiwari\}@iiti.ac.in, 
	sharmila@sit.ac.in  
}

\maketitle
\thispagestyle{firstpage} 
\begin{abstract}
	Multiclass classification is a fundamental problem across a wide range of domains. It is still challenging due to possession of high inter-class similarity, class imbalance datasets, and variability in data distributions. Rule-based classifiers such as XGBoost often achieve stronger performance on structured features, but they are limited in capturing smooth functional relationships among variables. Similarly, neural network models can represent complex nonlinear interactions but frequently suffer from overfitting and generalization issues. To address these limitations, we  propose LFS-FRAME, a leakage-free stacked ensemble framework that integrates functional learning using Kolmogorov-Arnold Networks (KAN) and rule-based learning via XGBoost for robust multiclass classification.
	The proposed framework constructs unbiased meta-features by employing a strict out-of-fold stacking strategy to ensure complete isolation between training and validation data hence  preventing performance leakage. By learning over probabilistic outputs from heterogeneous base learners, the meta-classifier effectively exploits both global functional patterns and sharp decision boundaries present in the complex data. Experimental evaluations on multi-class datasets demonstrate that LFS-FRAME improves  performance metrics, and overall accuracy is 89.85\% in identifying major families and 81.74\% in identifying sub-families relative to strong single-model baselines. These results highlight the effectiveness of leakage-free functional and rule-based stacking for reliable and generalizable multiclass classification.
	
\end{abstract}

\noindent 

\section{Introduction}
Ensemble learning has emerged as a cornerstone of modern machine learning, enabling the combination of multiple models to achieve superior predictive performance over individual learners\cite{Xue2026}.  Initially popularized through bagging (e.g., Random Forests) and boosting (e.g., AdaBoost, Gradient Boosting) in the late 1990s and early 2000s, ensemble methods gained prominence by exploiting the bias-variance tradeoff, reducing overfitting, and improving generalization on complex datasets.  Their rise accelerated with the availability of computational resources and large-scale data, leading to widespread adoption across domains like computer vision (e.g., deep ensemble CNNs)\cite{Wu2025}, 
 finance (risk modeling)\cite{Zhang2025,Aruleba2025,banking}, healthcare (diagnostic classifiers)\cite{Imrie2025}, and cybersecurity (malware detection)\cite{Moujoud2025,Vasan2025,ICOECA}. 

Among advanced ensemble techniques, stacking stands out for its meta-learning approach, where a second-level model learns to optimally blend predictions from diverse base classifiers. Wolpert\cite{Wolpert1992}  formalized stacking in 1992, which renewed interest not only in multiclass settings, but also intrusion detection and multi-label malware classification\cite{AugSPORF}, where it leverages class probability vectors from bases like RF, SVM, and XGBoost to boost accuracy by 5-15\% over single models. 
Despite these advantages, stacking ensembles, particularly for multiclass classification, face several critical challenges that can undermine their effectiveness if unaddressed.

\noindent
Key issues in Multi-class stacking ensembles:

Stacking introduces complexities amplified by the high-dimensional probability outputs (\(B \times C\) features) in multiclass problems.\\
\noindent
1. Data leakage:
Training the meta-learner on in-sample base predictions leaks label information, inflating performance metrics unrealistically. \\
\noindent
2. Meta-learner overfitting:
High-dimensional meta-features lead to overfitting unless regularized (e.g., via simple logistic regression). \\
\noindent
3. Base model diversity deficiency:
Correlated bases provide redundant signals, diminishing stacking gains. \\
\noindent
4. Computational overhead:
CV-based out-of-fold predictions scale poorly with \(B\), \(C\), and folds. \\
Apart from these the other issues include, class imbalance and interpretability, necessitate careful design, as outlined in prior analyses. 

To address these issues, in this paper we propose a leakage free multiclass stacked ensemble of functional and rule based models\footnote[1]{This paper is presented at \textbf{IEEE World Congress on Computational Intelligence (WCCI)} from 21–26 June 2026 Maastricht, 	The Netherlands}.  Unlike  conventional stacking ensemble the proposed framework is a structured integration of functional and rule-based learners, designed explicitly for leakage-free, probability-level multiclass classification. 
Its novelty lies in the principled fusion of complementary inductive biases and its dynamic adaptability to evolving multi-class families. 
\section{Background and Related Work}
\label{Background}
Multiclass classification (MCC) is a fundamental problem in machine learning, with wide-ranging applications in pattern recognition, healthcare analytics, bioinformatics, and security-oriented data analysis. As the number of classes increases, classifiers face significant challenges such as inter-class similarity, overlapping decision regions, and severe class imbalance, which often lead to unstable generalization and degraded performance. Traditional single-model approaches frequently struggle in such high-cardinality settings, motivating the adoption of ensemble learning techniques to improve robustness and predictive accuracy \cite{dietterich2000,rokach2010,OCIT,MaxVote}.

Tree-based ensemble methods, particularly gradient-boosted decision trees, have demonstrated strong empirical performance on structured and tabular datasets. XGBoost, in particular, has become a widely adopted model due to its ability to learn sparse, non-linear decision rules while incorporating regularization to mitigate overfitting \cite{chen2016}. Several studies have reported effectiveness of XGB in multiclass learning scenarios; however, in our proposed work, its limitations concerned with reliance on piecewise-constant approximations is counteracted by adopting functional learning.

Neural network–based models provide a complementary perspective by approximating complex functional mappings through continuous transformations. Deep learning approaches have been applied to multiclass classification with varying degrees of success, yet they often suffer from sensitivity to hyperparameter tuning, overfitting under limited data, and reduced interpretability \cite{LeCun2015}. Recently, Kolmogorov–Arnold Networks (KAN) have been proposed as a theoretically grounded functional learning paradigm based on the Kolmogorov–Arnold representation theorem, enabling multivariate function approximation through compositions of univariate nonlinear functions \cite{Liu2024}. KANs have also excelled for being resilient to generative adversarial attacks \cite{4L5FKAN,COMSNETS,RaaS}. While KANs offer improved parameter efficiency and interpretability, their standalone use may be insufficient for modeling sharp decision boundaries required in highly discriminative tasks.

Stacking generalization, originally introduced by Wolpert \cite{Wolpert1992}, aims to overcome the limitations of individual learners by training a meta-classifier on the predictions of multiple base models. Numerous stacking-based frameworks have been proposed, including homogeneous ensembles, heterogeneous learners, and probability-level fusion strategies \cite{Polikar2012}. 
Despite these, many existing stacking approaches inadvertently introduce information leakage, as the meta-classifier is often trained on predictions generated from the same data used to train the base models. Such leakage leads to overly optimistic performance estimates and poor real-world generalization, particularly in multiclass settings where error propagation is amplified \cite{Cawley2010}.

More recent studies have emphasized the importance of out-of-fold (OOF) prediction strategies to mitigate leakage and ensure unbiased meta-feature construction \cite{Bühlmann2003}. Separately, hybrid ensembles combining neural and tree-based models have shown potential benefits; however, these approaches often lack a principled integration strategy and theoretical grounding, and few explicitly address leakage prevention in a systematic manner.

\section{Problem Definition}
\label{sec:Problem}
\noindent
Definition: Leakage-Free Multiclass Stacking Framework\\
A Multiclass Stacking Framework built for a dataset \( D = {(x_i, y_i)}_{i=1}^{N} \), is said to be Leakage-Free if the meta-feature vector for sample \( x_i \) is constructed as:
\(
z_i = [\hat{p}^{(1)}_i, \hat{p}^{(2)}_i, \dots, \hat{p}^{(M)}_i]
\)
where each \( \hat{p}^{(m)}_i \) is the out-of-fold class probability vector produced by the \( m^{th} \) base learner and
no classifier is allowed to observe \( y_i \) while producing any component of \( z_i \).

Let the training set be \(\mathcal{D} = \{(x_i, y_i)\}_{i=1}^n	\)  
where \(x_i \in \mathcal{X}\) is a feature vector and \(y_i \in \{1,\dots,C\}\) is the true multiclass label (\(C\) classes). 
Suppose we have \(B\) base classifiers \(\{f_b\}_{b=1}^B\), each mapping  
\(	f_b : \mathcal{X} \to [0,1]^C	\)  
so \(f_b(x_i)\) is a probability vector over the \(C\) classes. 
The meta‑learner \(g\) is a classifier that takes as input the concatenated base‑model outputs:  
\(	z_i = \big[ f_1(x_i)^\top, f_2(x_i)^\top, \dots, f_B(x_i)^\top \big]^\top \in [0,1]^{B \times C}	\)  
and predicts the final class  
\(	\hat{y}_i = g(z_i).	\)  

In the leaky version, each base model \(f_b\) is trained on the full training set \(\mathcal{D}\):  
\(	f_b = \mathcal{A}_b(\mathcal{D})	\)  
where \(\mathcal{A}_b\) is the learning algorithm for base model \(b\). 
Then the meta‑level training set is built as  \(	\mathcal{Z} = \{(z_i, y_i)\}_{i=1}^n	\quad\text{where}\quad
	z_i = \big[ f_1(x_i)^\top, \dots, f_B(x_i)^\top \big]^\top.	\)  
The meta‑learner is trained as  
\(	g = \mathcal{A}_g(\mathcal{Z}).	\) 
Here the leakage is that \(f_b(x_i)\) is computed on a model that has already seen \((x_i, y_i)\) during training. That is, the label \(y_i\) has already influenced the learned parameters of \(f_b\), so the probability vector \(f_b(x_i)\) is correlated with \(y_i\) in a way that will not hold on new, unseen data. 
This means,	\(
	\text{Cov}(y_i, f_b(x_i)) \text{ is inflated on } \mathcal{D}	\),  
but this covariance will shrink on test data \(\mathcal{D}_{\text{test}}\), causing the meta‑learner to overfit to training‑specific patterns and leading to optimistic in‑sample performance. 

\section{Proposed Methodology}
\label{sec:Proposed}
To remove leakage, we have employed out‑of‑fold (OOF) predictions described as follows: 
By partitioning \(\mathcal{D}\) into \(K\) folds \(\{\mathcal{D}_k\}_{k=1}^K\). 
For each fold \(k\) we perform the following:\\  
- Train each base model \(f_b^{(k)}\) on \(\mathcal{D} \setminus \mathcal{D}_k\):  
	$
	f_b^{(k)} = \mathcal{A}_b(\mathcal{D} \setminus \mathcal{D}_k).
	$ \\ 
- Compute predictions only on \(\mathcal{D}_k\)  \\
	$
	z_i^{(k)} = \big[ f_1^{(k)}(x_i)^\top, \dots, f_B^{(k)}(x_i)^\top \big]^\top,
	\quad (x_i, y_i) \in \mathcal{D}_k.
	$\\
\noindent	  
- Concatenate all OOF predictions into the meta‑level training set:  
\(	\mathcal{Z}_{\text{oof}} = \{(z_i^{(k)}, y_i)\}_{(x_i,y_i)\in\mathcal{D}_k,\, k=1}^K.	\) \\
\noindent 
- Train the meta‑learner as  
\(	g = \mathcal{A}_g(\mathcal{Z}_{\text{oof}}).\)  
	
Further, for every sample \(x_i\), the vector \(z_i^{(k)}\) is generated by base models \(f_b^{(k)}\) that never saw \((x_i, y_i)\) during training, so  
	\(
	f_b^{(k)}(x_i) \perp y_i \mid \mathcal{D} \setminus \mathcal{D}_k
	\)  
	in the sense that the label information is not directly leaked into the meta‑level features. \\
At the test time, for a new and unseen sample \(x^*\) we perform the following:\\  
- Retrain each base model on the full \(\mathcal{D}\):  
	\(
	f_b^{\text{full}} = \mathcal{A}_b(\mathcal{D}).
	\)  \\
- Compute the meta‑features: \\ 
	\(
	z^* = \big[ f_1^{\text{full}}(x^*)^\top, \dots, f_B^{\text{full}}(x^*)^\top \big]^\top.
	\)  \\
- Predict:  
	\(
	\hat{y}^* = g(z^*).
	\)  
	
Because \(g\) was trained only on OOF predictions that did not leak labels, and the test predictions \(f_b^{\text{full}}(x^*)\) are generated on unseen data, there is no label leakage into the meta‑level model in our proposed approach, and the performance estimate generalizes more reliably. 

\begin{figure}[t]
	\centering
	\includegraphics[width=6cm,height=9cm, scale=0.4]{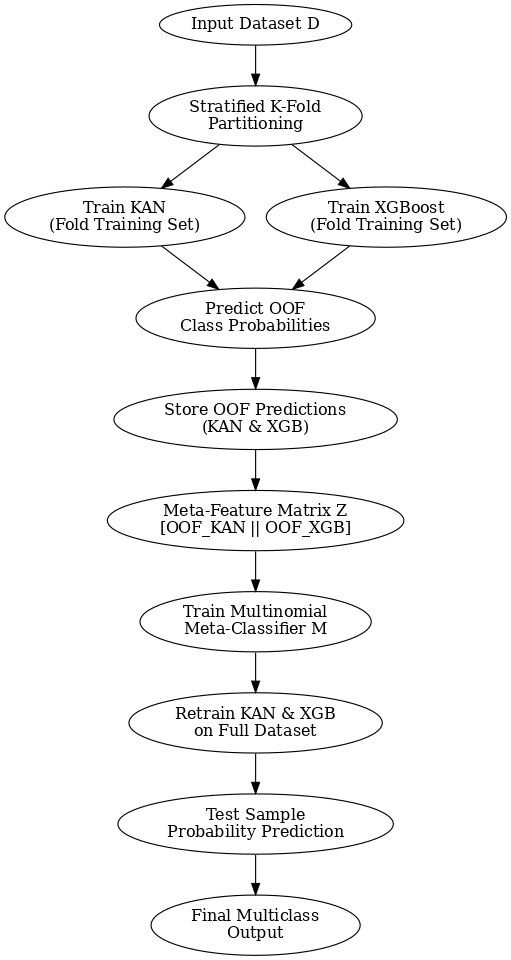}
	\caption{Leakage-free multiclass stacking framework integrating Kolmogorov--Arnold Networks (KAN) and XGBoost using out-of-fold probability fusion.}
	\label{fig:kan_xgb_stacking}
\end{figure}

In Algorithm \ref{Algo1} we present a leakage-free multiclass stacking framework that integrates KAN and XGBoost as complementary base learners. Reason for choosing this combination is already presented in the section \ref{Background}. The dataset is first partitioned using stratified K-fold cross-validation to preserve class distributions across folds. For each fold, KAN and XGBoost models are trained exclusively on the fold-specific training subset, and class probability predictions are generated only for the corresponding validation subset. These out-of-fold (OOF) probability estimates are stored and later concatenated to form a meta-feature matrix, ensuring that the meta-classifier is trained on predictions obtained from models that have never seen the same samples during training, thereby eliminating information leakage. A multinomial meta-classifier is then trained on these OOF features to learn optimal fusion weights across base models. Finally, KAN and XGBoost are retrained on the full dataset, and their probability outputs for unseen test samples are fused through the trained meta-classifier to produce the final multiclass prediction. This design ensures robust generalization, unbiased performance estimation, and effective utilization of functional (KAN) and rule-based (XGBoost) decision paradigms. Figure \ref{fig:kan_xgb_stacking} further facilitates to enhance the clarity of the proposed framework.	

The proposed method resolves key gaps in multiclass malware classification by combining leakage-free stacking, probability-level fusion, and complementary functional–rule learners within a scalable and robust ensemble framework.

\begin{table}
	\caption{Details of existing works comparison with evaluated dataset}
	\label{tab:BaselineComparison}
	\centering
	\begin{tabular}{|c|c|c|c|c|l|}
		\hline
		Sl.&Reference &Methodology used	 & MCC(Accuracy \%)  \\ \hline
		1.&\cite{roy} &HyStack Ensemble	 & 4CC: 85.04 16CC:70.29  \\ \hline
		2.&\cite{shafin} &Hybrid CNN BiLSTM	 & 3CC: 84.54 15CC:72.6  \\ \hline
		3.&\cite{cevallos} &SMOTE DNN	 & 4CC: 75.4 16CC:68.2   \\ \hline
		4.&\cite{sharmila} &RF with Hyp Tuning	 & 4CC: 89.07 16CC:68.2   \\ \hline
		5.&\textbf{Ours} &\textbf{KAN+XGB}	 & \textbf{4CC: 89.85 16CC:81.74 }  \\ \hline
		%
		%
		%
		%
	\end{tabular}
\end{table}
\begin{algorithm}[H]
	\begin{algorithmic}[1]
		\caption{\textbf{Leakage-Free Multiclass Stacking with KAN and XGBoost}}
		\label{Algo1}
		\renewcommand{\algorithmicrequire}{\textbf{Input:}}
		\renewcommand{\algorithmicensure}{\textbf{Output:}}
			
		\REQUIRE $D = \{(x_i, y_i)\}_{i=1}^{N}$, number of folds $K$
		\ENSURE  Trained meta-classifier $M$		
		\\ \textit{Initialisation} :
		\STATE Initialize StratifiedKFold with $K$ folds
		\STATE Initialize out-of-fold probability matrices:
		\STATE \hspace{0.5cm} $\text{OOF}_{KAN} \in \mathbb{R}^{N \times C}$,
		$\text{OOF}_{XGB} \in \mathbb{R}^{N \times C}$
	
		\STATE For{$f = 1$ to $K$}
		\STATE \hspace{0.2cm}Split $D$ into training set $D_{\text{train}}^{(f)}$ and validation set $D_{\text{val}}^{(f)}$
				
		\STATE \hspace{0.2cm}Train Kolmogorov--Arnold Network $KAN^{(f)}$ on $D_{\text{train}}^{(f)}$
		\STATE \hspace{0.2cm}Train XGBoost classifier $XGB^{(f)}$ on $D_{\text{train}}^{(f)}$
				
		\STATE \hspace{0.2cm}Obtain class probability predictions on $D_{\text{val}}^{(f)}$:
		\STATE \hspace{0.5cm} $P_{KAN} \leftarrow KAN^{(f)}.\text{predict\_proba}(D_{\text{val}}^{(f)})$
		\STATE \hspace{0.5cm} $P_{XGB} \leftarrow XGB^{(f)}.\text{predict\_proba}(D_{\text{val}}^{(f)})$
				
		\STATE \hspace{0.2cm}Store out-of-fold predictions:
		\STATE \hspace{0.5cm} $\text{OOF}_{KAN}[\text{val\_idx}] \leftarrow P_{KAN}$
		\STATE \hspace{0.5cm} $\text{OOF}_{XGB}[\text{val\_idx}] \leftarrow P_{XGB}$
		\STATE EndFor
		\STATE Construct meta-feature matrix:
		\STATE \hspace{0.5cm} $Z \leftarrow [\text{OOF}_{KAN} \;||\; \text{OOF}_{XGB}]$
		\STATE Train multinomial meta-classifier $M$ on $(Z, y)$
		\STATE Retrain $KAN$ and $XGB$ on the full dataset $D$
		\STATE For each{ test sample $x_{\text{test}}$}
		\STATE \hspace{0.2cm}Obtain probability vectors:
		\STATE \hspace{0.5cm} $p_{KAN} \leftarrow KAN.\text{predict\_proba}(x_{\text{test}})$
		\STATE \hspace{0.5cm} $p_{XGB} \leftarrow XGB.\text{predict\_proba}(x_{\text{test}})$
		\STATE \hspace{0.2cm}Concatenate probabilities:
		\STATE \hspace{0.5cm} $z_{\text{test}} \leftarrow [p_{KAN} \;||\; p_{XGB}]$
		\STATE \hspace{0.2cm}Predict final class label:
		\STATE \hspace{0.5cm} $\hat{y} \leftarrow M(z_{\text{test}})$
		\STATE EndFor
		\STATE Return meta-classifier $M$ output
	\end{algorithmic} 
\end{algorithm}

\begin{table*}[h]
	\centering
	\caption{Comparison Between Traditional Stacking and the Proposed Leakage-Free Stacked Ensemble}
	\label{tab:stacking_comparison}
	\renewcommand{\arraystretch}{1.25}
	\begin{tabular}{|p{4cm}| p{5.5cm}| p{6cm}|}
		\hline
		\textbf{Aspect} & \textbf{Traditional Stacking} & \textbf{Proposed Leakage-Free Stacking (LFS-FRAME)} \\
		\hline
		Meta-feature construction 
		& Base-model predictions often generated on training data 
		& Strictly uses \textbf{out-of-fold (OOF)} predictions for meta-feature construction \\ \hline
		
		Information leakage 
		& High risk due to reuse of training samples across levels 
		& \textbf{Explicitly eliminated} through fold-wise data isolation \\ \hline
		
		Generalization reliability 
		& Validation and test performance may be optimistically biased 
		& \textbf{Unbiased and stable} generalization estimates \\ \hline
		
		Base learner diversity 
		& Typically homogeneous or weakly heterogeneous models 
		& \textbf{Heterogeneous ensemble}: functional (KAN) + rule-based (XGBoost) \\ \hline
		
		Error correlation 
		& High due to similar inductive biases 
		& \textbf{Reduced} via complementary learning mechanisms \\ \hline
		
		Multiclass scalability 
		& Performance degrades as number of classes increases 
		& \textbf{Maintains macro-level performance} in high-class regimes \\ \hline
		
		Theoretical grounding 
		& Primarily empirical with limited formal justification 
		& \textbf{Grounded in functional approximation and ensemble theory} \\ \hline
		
		Robustness to class imbalance 
		& Often biased toward majority classes 
		& \textbf{Class-weighted learning and probability-level fusion} \\ \hline
		
		Interpretability 
		& Limited or inconsistent 
		& Rule-based component retains interpretability \\ \hline
		
		Extensibility 
		& Pipeline redesign required to add new models 
		& \textbf{Modular and easily extensible} stacking architecture \\
		\hline
	\end{tabular}
\end{table*}

\begin{table}[h]
	\centering
	\caption{KAN and XGB Hyperparameters and values }
	\label{tab:KAN+XGBhyp}
	\renewcommand{\arraystretch}{1.25}
	\begin{tabular}{|p{2.8cm}| p{1cm}| p{2.8cm}| p{0.8cm}| }
		\hline
		\textbf{KAN-Hyperparameter} & \textbf{Value}& \textbf{XGB-Hyperparameter} & \textbf{Value}  \\
		\hline
		 hidden\_layer\_size     & 512   &eta (learning rate)     & 0.1  \\ 
		 \hline
		regularize\_activation & 0.0 &max\_depth  & 6     \\
		\hline
		 regularize\_entropy    & 0.0 &subsample    & 0.8    \\
		\hline
		 regularize\_ridge      & 0.2 &colsample\_bytree      & 0.8    \\
		\hline
		 spline\_order          & 3  &seed          & 42      \\
		\hline
		 batch\_size            & 32 &-&-     \\
		\hline
		 learning\_rate (lr)    & 0.0025 &-&-\\
		\hline
		 weight\_decay          & 0.0001 &-&-\\
		\hline
		 n\_epochs              & 100  &-&-  \\
		\hline
			
	\end{tabular}
\end{table}


\begin{table}[h]
	\centering
	\caption{Classification performance accuracy of XGB, KAN and KAN+XGB for both datasets - 16CC}
	\label{tab:Perf}
	\renewcommand{\arraystretch}{1.25}
	\begin{tabular}{|p{2cm}|p{1.5cm}| p{1.5cm}| p{1.5cm}|}
		\hline
		\textbf{Dataset} & \textbf{KAN-only} & \textbf{XGB-only} & \textbf{KAN+XGB}  \\
		\hline
		CIC    & 69.66 & 76.21 & 79.32   \\
		\hline
		EnhancedCIC  & 71.82 & 79.32 & 81.74   \\
		\hline
	\end{tabular}
\end{table}

\section{ Experimental Details with Results and discussion}
To evaluate the proposed model we have chosen publicly available dataset having 16 classes and its enhanced version containing more samples with adversarial training. This section provides detailed dataset description followed by the chosen functional and rule-based learning models. We also provide the baseline comparison alongwith the advantages of the proposed framework. The sample source segment is available at our repository https://github.com/sharmilaharsha/Leakfree-WCCI2026.
\subsection{ Dataset Description}
We have employed two datasets in our work: the public volatile-memory dataset CIC-MalMem-2022 
(malware memory analysis)\cite{CIC2022} and a larger internal dataset called EnhancedVolMem dataset 
created from CIC-MalMem-2022 via a set of augmentation and synthetic-sample generation
techniques (GAN, interpolation, controlled mixing). Further, we summarize contents,
provenance and the role each dataset plays in our experiments.\\
\textbf{CIC-MalMem-2022}: 
Obfuscated malware is a malware that hides to avoid detection and
extermination. The obfuscated malware dataset is designed to test obfuscated malware detection methods through memory. The dataset was created to represent as close to a real-world situation as possible using malware that is prevalent in the real world. It is composed of Spyware, Ransomware and Trojan malware, and a balanced dataset that can be used to test obfuscated malware detection systems.
This dataset uses debug mode for the memory dump process to avoid the dumping
process to show up in the memory dumps. This works to represent a more accurate
example of what an average user would have running at the time of a malware attack.\\
\textbf{EnhancedCIC dataset: } 
This is a larger dataset produced by augmenting CICMalMem-2022 using a combination of controlled synthetic-sample generation and
interpolation techniques. Methods used include generative adversarial network (GAN)
based synthesis, feature-space interpolation (SMOTE-like and linear interpolation between pairs of feature vectors), and conservative class-balanced oversampling. The goal is to increase per-class sample counts so that multi-class attribution experiments (e.g., 15/16-class tasks) have sufficient data.
The train, test and validation split chosen with these datasets is 70:15:15. Sample dataset is available in our GitHub repository.

\begin{figure}[t]
	\centering
	\includegraphics[width=9cm,height=6cm, scale=0.4]{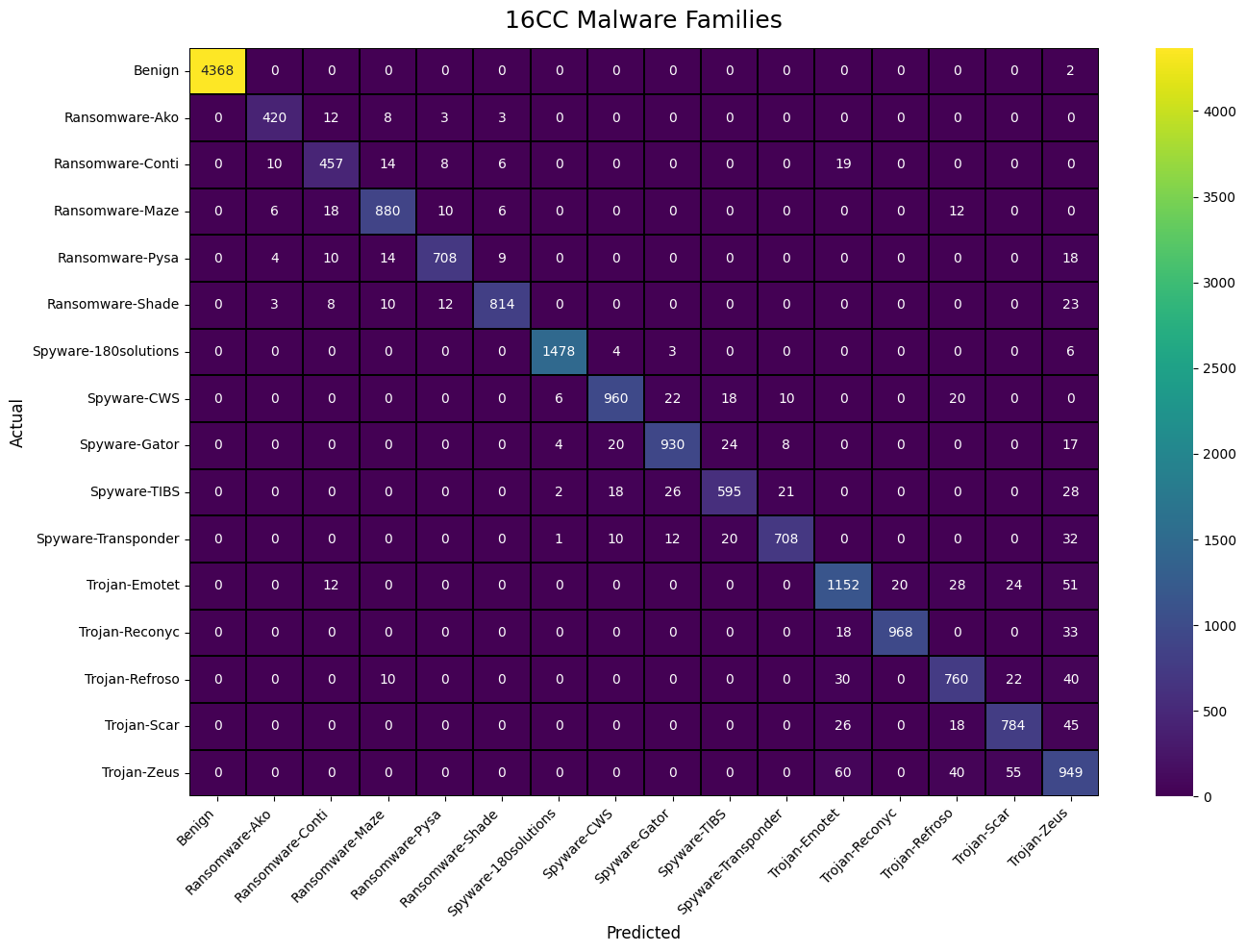}
	\caption{Confusion Matrix derived from KAN+XGB with OOF for 16CC.}
	\label{fig:CM16cc}
\end{figure}

\begin{figure}[h]
		\centering
		\includegraphics[width=9cm,height=5cm, scale=0.4]{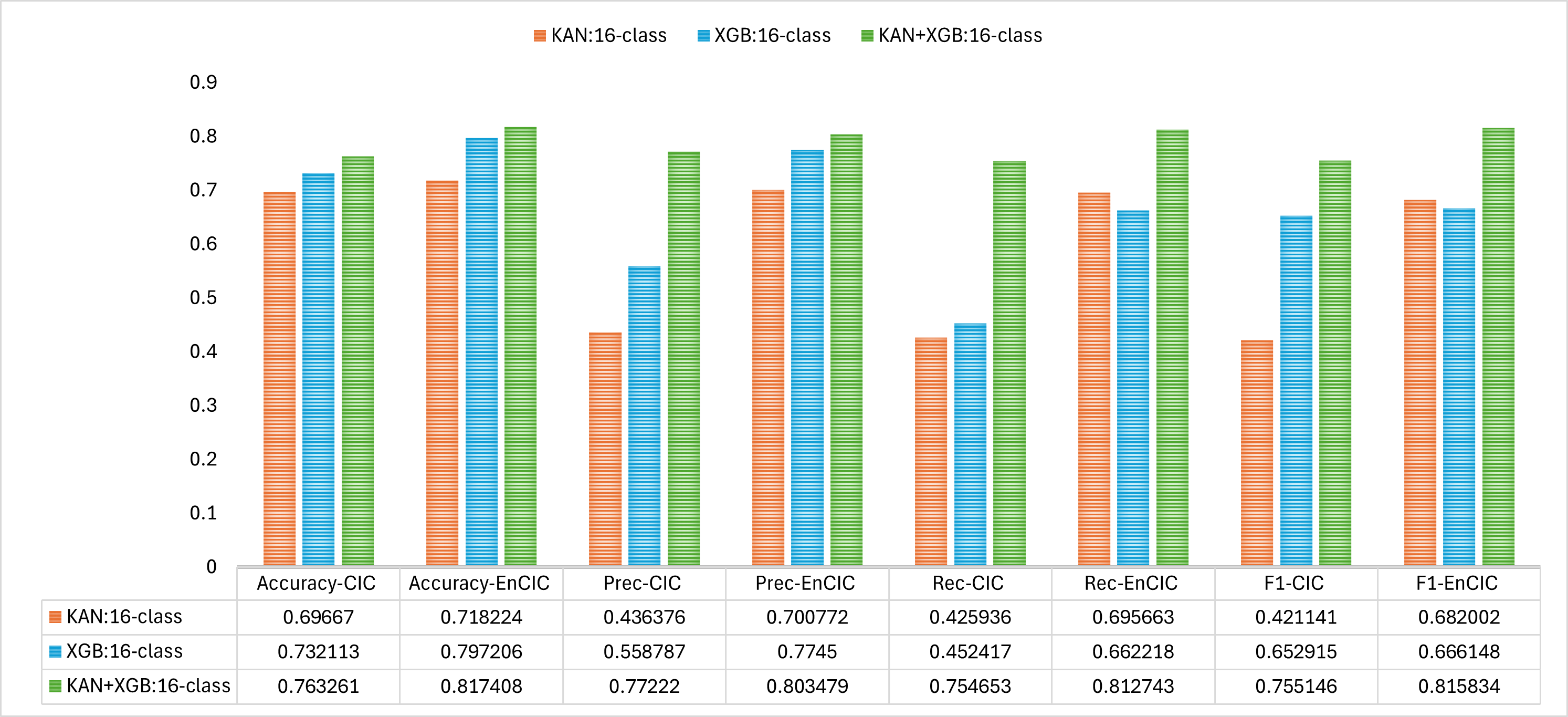}
		\caption{KAN-XGB-16CC.}
		\label{fig:KAN-XGB-16cc}
\end{figure}

\subsection{  Learning Models}
In this section we describe the functional learner(KAN), rule-based learners(XGB) and the combination with OOF.\\
\textbf{Kolmogorov–Arnold Network (KAN):}\\
The Kolmogorov–Arnold Network (KAN) serves as a functional learner that models smooth multivariate mappings \( f: \mathbb{R}^d \rightarrow \mathbb{R}^C \) using compositions of univariate nonlinear basis functions, as motivated by the Kolmogorov–Arnold representation theorem. Given an input feature vector \( \mathbf{x} \in \mathbb{R}^d \), KAN approximates class logits \( \mathbf{z} = f(\mathbf{x}) \), which are transformed into class probabilities via a softmax function. This formulation enables KAN to capture global, continuous feature interactions efficiently; however, its reliance on smooth approximations can limit its ability to model sharp class boundaries when used in isolation.

\noindent
\textbf{Extreme Gradient Boosting (XGBoost):}\\
XGBoost is employed as a rule-based learner that models class posterior probabilities through an additive ensemble of decision trees, expressed as \( \hat{y} = \sum_{m=1}^{M} f_m(\mathbf{x}) \), where each \( f_m \) represents a regression tree optimized via gradient boosting. In the multiclass setting, a softmax objective is used to estimate \( P(y=c \mid \mathbf{x}) \) for each class \( c \). XGBoost excels at capturing localized, non-linear decision regions and sparse feature interactions. 

\noindent
\textbf{KAN+XGB Out-of-Fold Stacked Ensemble:}\\
The proposed ensemble integrates KAN and XGBoost using a leakage-free out-of-fold (OOF) stacking strategy. For each fold \( k \), base learners are trained on \( \mathcal{D}*{\text{train}}^{(k)} \) and generate unbiased probability estimates \( \mathbf{p}^{(k)}*{\text{KAN}} \) and \( \mathbf{p}^{(k)}*{\text{XGB}} \) for the held-out subset \( \mathcal{D}*{\text{val}}^{(k)} \). These predictions are concatenated to form meta-features \( \mathbf{z} = [\mathbf{p}*{\text{KAN}} | \mathbf{p}*{\text{XGB}}] \), which are used to train a multinomial meta-classifier. This formulation exploits complementary functional and rule-based representations while preventing information leakage, leading to improved multiclass generalization.
Hyperparameters tuning details are tabulated in Table \ref{tab:KAN+XGBhyp}. 

\subsection{Baseline Comparison}
The comparative results are summarized in Table \ref{tab:BaselineComparison} for CIC dataset.  It highlights clear differences between existing approaches and our approach. Earlier ensemble-based methods such as HyStack\cite{roy}  achieved reasonable performance for the 4CC showing 85.04\% accuracy, but there is a  substantial drop when extended to 16CC classes (showing 70.29\%). This indicates limited robustness of this model for fine-grained classes. Similarly, deep learning–based approaches, including the Hybrid CNN–BiLSTM \cite{shafin} and SMOTE-DNN \cite{cevallos}, showed competitive results in 4CC (with 84.54\% and 75.4\% respectively). Yet they suffer from degraded performance under higher number of class cases, with accuracies falling to the lower 60\% to 70\% for higher number of classes. This suggests that, representation learning and data balancing help in simpler scenarios, but these methods struggle to preserve discriminative capability as number of classes increases.

Although, the Random Forest with hyperparameter tuning \cite{sharmila} improves 4CC performance to 89.07\%, demonstrating the strength of rule-based learners on structured features. However, there is a sharp decline to 68.2\% in the 16CC, it reveals the limitations in generalization when decision boundaries become more complex and overlapping. In contrast to this, our proposed KAN+XGB framework with OOF consistently outperforms all compared methods across both settings, achieving 89.85\% accuracy for 4CC and a markedly higher 81.74\% for 16CC. This improvement in the high-class regime is particularly significant. Thus it  demonstrates the ability of the proposed leakage-free stacked ensemble for class-wise discrimination even under increased multiclass complexity. Fig. \ref{fig:CM16cc} depicts the confusion matrix. 
Finally, Fig.\ref{fig:KAN-XGB-16cc} portrays the effect of stacking over individual learning methods.
Table \ref{tab:Perf} presents classification performance accuracy of XGB, KAN and KAN+XGB for both datasets evaluated for 16 class classification. 
Overall, the comparison indicates that the proposed approach not only exceeds state-of-the-art performance in simpler classification scenarios but, also delivers substantial gains in challenging multiclass scenarios. The combination of functional learning (KAN) and rule-based modeling (XGBoost), coupled with leakage-free stacking, enables better robust and consistent performance existing methods.
\subsection{ Advantages of the LFS-FRAME}
The proposed leakage-free stacked ensemble offers a robust solution for multiclass classification by integrating complementary functional and rule-based learning paradigms within a rigorously unbiased stacking framework. By combining KAN, which effectively model smooth and compositional functional relationships, with XGB, which excels at learning sharp, rule-based decision boundaries, the approach expands the hypothesis space and improves class discrimination. The use of strict out-of-fold stacking ensures complete isolation between training and validation data, eliminating information leakage and enabling reliable generalization. 

\subsection{Computational Trade-off}
\label{Computational Trade-off}
Alongwith the comparison between traditional stacking and leakage free stacking presented in Table \ref{tab:stacking_comparison}, there exists computational trade-offs. Unlike traditional leaky stacking approaches, while leakage-free stacking introduces a linear computational overhead due to out-of-fold training. This cost is both bounded and necessary to ensure unbiased meta-learning, making the resulting performance improvements statistically valid and practically reliable.
In traditional stacking especially for malware family identification, base models trained on the full dataset generate meta-features that inevitably leak training data information to the meta-learner, resulting in weak meta-learning validity, unstable generalization, and low reproducibility despite lower training costs, critical flaws when classifying evasive malware families, 
where over-optimistic performance (e.g., inflated accuracies from leaked biases) fails against real-world variants and concept drift. Leakage-free stacking, using out-of-fold predictions from k-fold cross-validation, eliminates this information leakage by ensuring the meta-learner trains solely on unbiased hold-out predictions, yielding asymptotically equivalent performance with marginally higher $O(k)$ training overhead but delivering strong meta-learning validity, robust generalization reliability, and high review defensibility. This approach is particularly compelling for malware family detection 
that reliably identify families in dynamic, adversarial environments unlike biased traditional methods.
\section{Conclusion}
\label{Conclusion}
This paper presented LFS-FRAME, a leakage-free stacked ensemble framework. It is designed to address the inherent challenges in multiclass classification. 
The proposed approach leverages complementary inductive biases that are difficult to capture using a single modeling paradigm by integrating functional learning through Kolmogorov-Arnold Networks (KAN) with rule-based learning using XGBoost. To ensure unbiased meta-feature construction a strict out-of-fold stacking strategy is employed to eliminate information leakage resulting in reliable performance estimation.
Based on our experimental evaluation across multiple classes of CIC dataset, LFS-FRAME demonstrated excellence against existing ensemble and deep learning–based approaches, particularly in higher multiclass classification. While several baseline methods exhibited competitive accuracy in low-class scenarios, their performance degraded substantially as the number of classes increased. In contrast to this, LFS-FRAME maintained strong metrics,  highlighting its robustness and scalability. These results infer that combining functional and rule-based models within a leakage-free stacking architecture provides a stable decision mechanism for complex multiclass problems.
In summary, this study establishes leakage-free functional and rule-based stacking as an effective and generalizable strategy for multiclass classification. As a future work this model can be explored for evaluation of other evolving datasets and various ablation studies with advanced statistical significant testing. 

\end{document}